\documentclass[conference]{IEEEtran}
\IEEEoverridecommandlockouts
\usepackage{textcomp}
\usepackage{siunitx}
\usepackage{graphicx, subfig}
\usepackage{caption}
\usepackage{multirow}
\usepackage{booktabs}
\DeclareCaptionLabelSeparator{colon}{.  }

\begin{document}
\title{Real-Time Wearable Gait Phase Segmentation For Running And Walking
\thanks{This work was supported in part by the Ministry of Science and Technology, Taiwan, under Grant 109-2634-F-009-022.}
\thanks{J. -D. Sui, W. -H. Chen, T. -Y. Shiang and T. -S. Chang, "Real-Time Wearable Gait Phase Segmentation for Running And Walking," 2020 IEEE International Symposium on Circuits and Systems (ISCAS), 2020, pp. 1-5, doi: 10.1109/ISCAS45731.2020.9181210.}
}




\DeclareRobustCommand*{\IEEEauthorrefmark}[1]{%
  \raisebox{0pt}[0pt][0pt]{\textsuperscript{\footnotesize\ensuremath{#1}}}}

\author{\IEEEauthorblockN{Jien-De Sui\IEEEauthorrefmark{1},
Wei-Han Chen\IEEEauthorrefmark{2}, Tzyy-Yuang Shiang\IEEEauthorrefmark{3} and
Tian-Sheuan Chang\IEEEauthorrefmark{4}}
\IEEEauthorblockA{\IEEEauthorrefmark{1}\IEEEauthorrefmark{4}Instituter of Electronics,
National Chiao Tung University, Hsinchu, Taiwan\\
\IEEEauthorrefmark{2}\IEEEauthorrefmark{3}Department of Athletic Performance,
National Taiwan Normal University, Taipei, Taiwan\\
Email: \IEEEauthorrefmark{1}vigia117.ee06g@nctu.edu.tw,
\IEEEauthorrefmark{2}gn01800083@gmail.com,
\IEEEauthorrefmark{3}tyshiang@gmail.com,
\IEEEauthorrefmark{4}tschang@g2.nctu.edu.tw}
}
\maketitle






\begin{abstract}
Previous gait phase detection as convolutional neural network (CNN) based classification task requires cumbersome manual setting of time delay or heavy overlapped sliding windows to accurately classify each phase under different test cases, which is not suitable for streaming Inertial-Measurement-Unit (IMU) sensor data and fails to adapt to different scenarios. This paper presents a segmentation based gait phase detection with only a single six-axis IMU sensor, which can easily adapt to both walking and running at various speeds. The proposed segmentation uses CNN with gait phase aware receptive field setting and IMU oriented processing order, which can fit to high sampling rate of IMU up to 1000Hz for high accuracy and low sampling rate down to 20Hz for real time calculation. The proposed model on the 20Hz sampling rate data can achieve average error of 8.86 ms in swing time, 9.12 ms in stance time and 96.44\% accuracy of gait phase detection and 99.97\% accuracy of stride detection. Its real-time implementation on mobile phone only takes 36 ms for 1 second length of sensor data.
\end{abstract}

\IEEEpeerreviewmaketitle

\section{Introduction}
Gait phase is an important parameter in a lot of applications such as personal health care [1-3] or sports training to further develop other gait parameters and applications such as gait cycle, stride length, stride height [4-5] and inertial navigation system (INS) [6]. To identify gait phase, optical or force  plates  based  methods can provide accurate results but suffer from high product cost and limited test environments. On the other hand, the IMU-based system is not only low cost but also unconstrained to specific environments.  \par

Based on IMU sensor data, previous works assume zero velocity and accelerations in the stance phase, denoted as ZVU  (zero  velocity  update)  [7-10], to classify gait phase. [11] developed a threshold-based method for gait classification with a gyroscope. [12] uses the dynamic threshold to detect the gait phase. [13] develops a gait event detection with a hidden Markov  Model  filter
.  However, these  algorithms  need  personalized or per sensor calibrations, and threshold setting to get accurate results, or are limited to the normal walking speed case. \par

Beyond above signal processing approaches, one popular way in recent years is to use neural network (NN), CNN or recurrent neural network (RNN) to classify gait phase since one stride can be divided into two phase: the stance phase, and the swing phase.\par
\begin{figure}[t]
\centering
    \includegraphics[width=7cm]{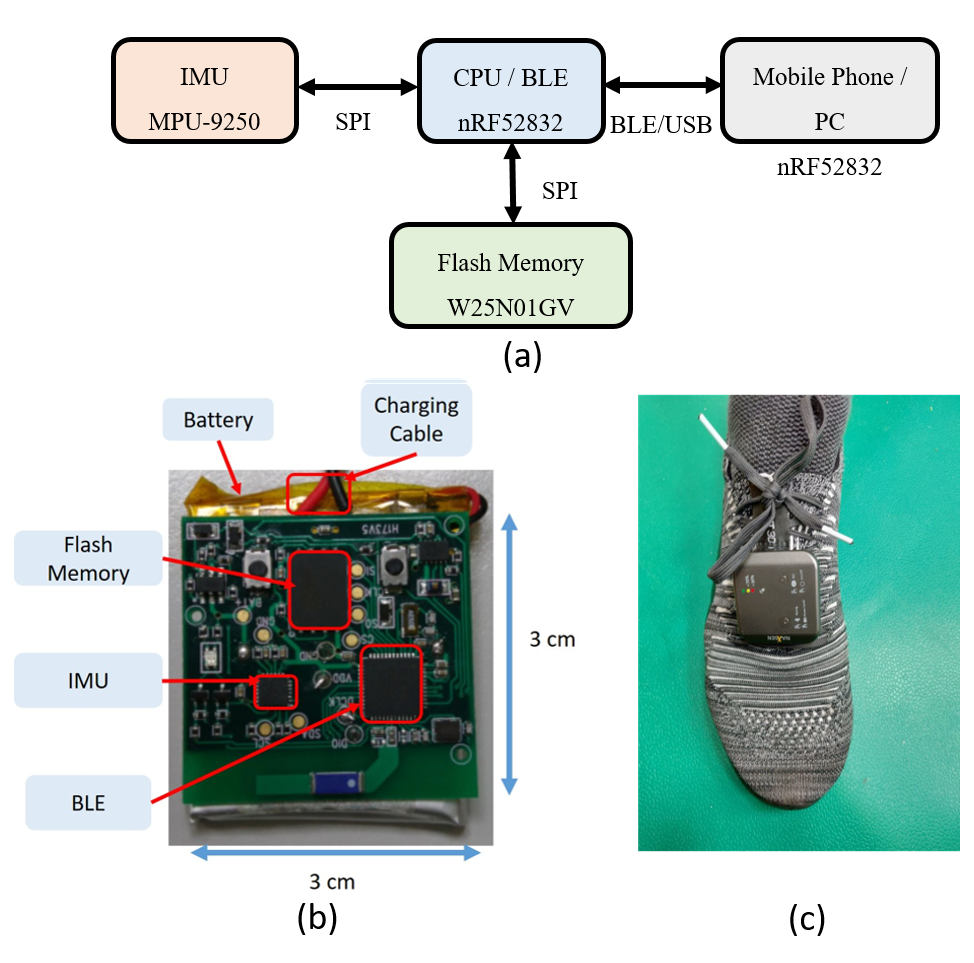}
    \caption{The sensor system architecture (a) and its PCB (b). (c) The position of sensor on the foot.} 
    \label{fig.Sensor}
\end{figure}

The NN, CNN or RNN based gait phase detection works [14-17] use the superior feature extraction capability of neural network to classify phase accurately. Their approaches consider this problem as a classification task. However, since the IMU sensor data is a time streaming data but NN requires a fixed size input, they split the input into segments, and classify each segment into one type of gait phase. Such whole segment classification is coarse grained and thus needs a sliding window approach or adjusts input delay to identify suitable start point of gait phase, which heavily depends on personal walking speed and types, and thus fails to extend to the running case. \par

To solve above problems, we treat this task as a segmentation problem with CNN. For this task, the segmentation approach enables per sample point classification instead of whole segment classification. This fine grained approach avoids the problem of personalized adjustment and is suitable for streaming sensor data and walking/running types. The CNN approach allows end-to-end training without explicit calibration, or ZVU assumption. The proposed segmentation network builds upon \emph{U-Net} [18], called \emph{IMU-Net}, but with gait phase aware receptive field setting and IMU oriented processing order to attain high accuracy. This lightweight network can run on mobile phones in real time to enable wearable applications.\par

\section{METHODOLOGY}
\subsection{Sensor}
Fig. \ref{fig.Sensor} shows the sensor system architecture, which consists of a 9-axis IMU from InvenSense (MPU-9250), 1Gb flash memory for storage, and a Bluetooth chip for wireless data transmission. The IMU consists of tri-axis accelerometer, tri- axis gyroscope and tri-axis magnetometer, but this paper only uses tri-axis accelerometer and tri-axis gyroscope. This sys- tem is integrated in a PCB of form factor 30 × 30 mm$^{2}$. This prototype consumes 9.3mA current for the highest sampling rate at 1000 Hz, and can ensure all day long continuous data recording with the 450 mAh rechargeable 3.7 V Li-ion battery. The sampling rate will be constrained to 20 Hz in case of real time data transmission to mobile phone due to speed constraint of Bluetooth.\par
The sensor is mounted and fixed on the shoe as shown in Fig. \ref{fig.Sensor} (c). The accelerometer has a dynamic range of ± 16 g (g = 9.81 m/s$^{2}$) and the gyroscope has a dynamic range of ± 2000 °/s [19].

\subsection{Data Collection}
Three normal subjects (two males, one female) with ages from 20 to 30 years old are chosen for the experiments.  The test tasks includes running and walking with various speeds (5, 7, 9, 11, 13, 17 and 19 km/h) based on the treadmill record. In addition to the speed, the data involves two common types of the strike: fore-foot strike (FFS) and rear-foot strikes (RFS). The ground truth data is collected by an optical measurement system, the Optojump system, which consists of transmitting and receiving bars to measure contact times during the activity with 1ms accuracy. These experiments are conducted on the treadmill and the flat ground. The amount of the whole raw dataset is 1593 seconds.

\subsection{Data preprocessing}
\begin{figure}[b]
    \includegraphics[width=\linewidth]{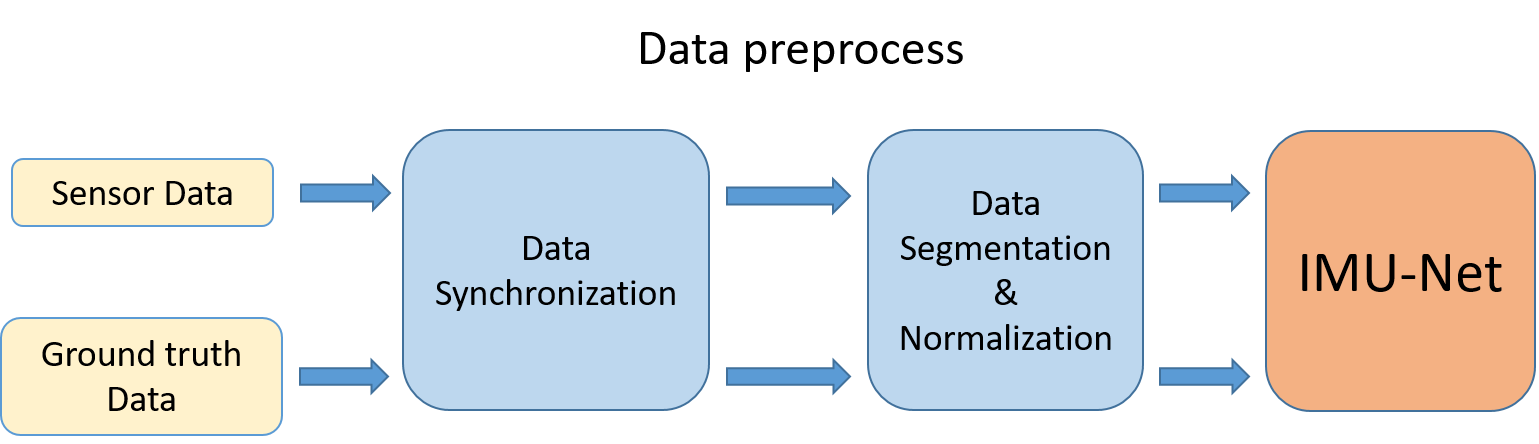}
    \caption{The whole processing pipeline diagram.} 
    \label{fig.preprocessflow}
\end{figure}

\begin{figure}[tb]
    \includegraphics[width=\linewidth]{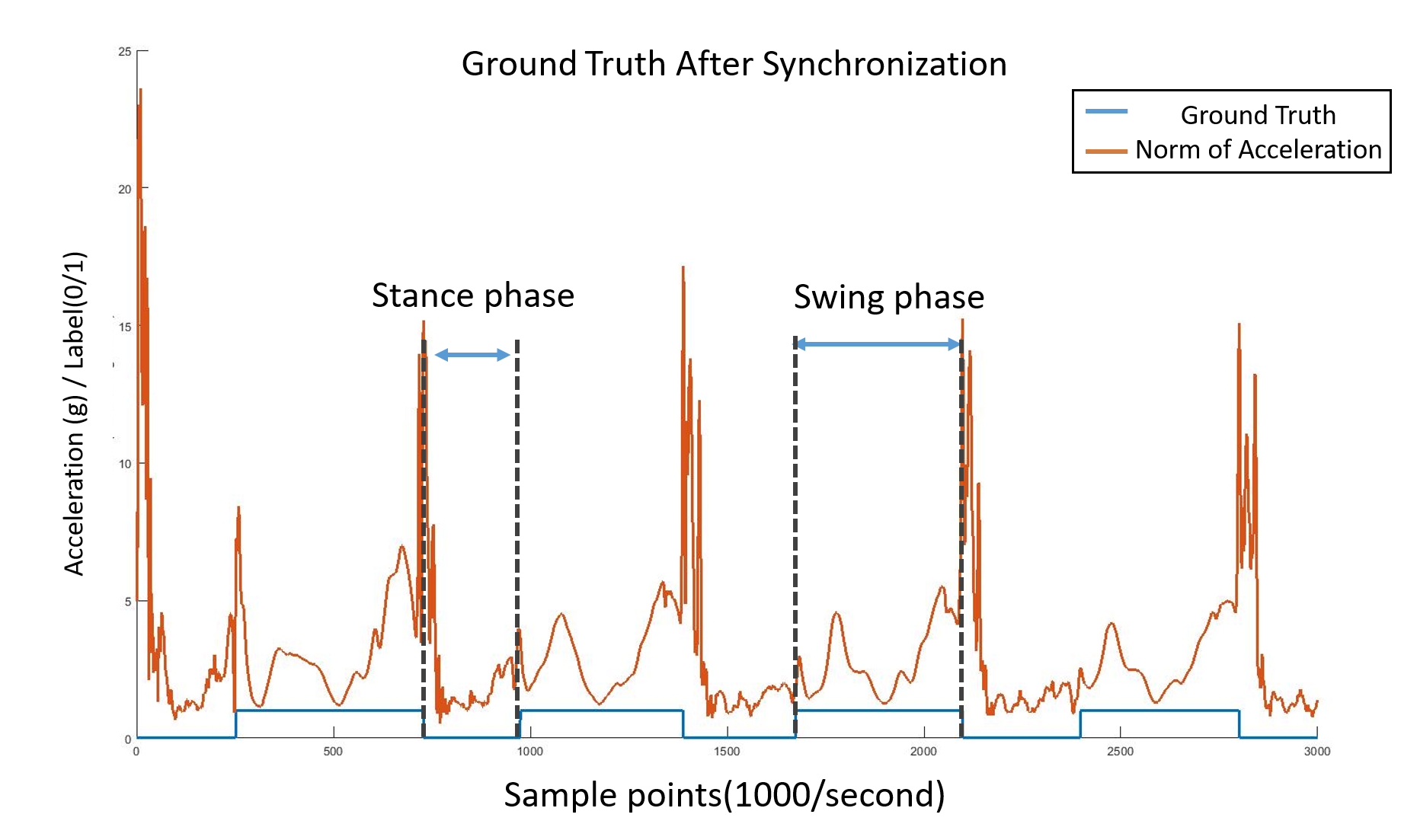}
    \caption{The data after synchronization step, where the blue line is ground truth label (0: stance, 1: swing), and orange line is the norm of tri-accelerometer. } 
    \label{fig.Aftersyn}
\end{figure}

Fig. \ref{fig.preprocessflow} shows the whole processing pipeline diagram, which consists of two data preprocessing steps before feeding to \emph{IMU-Net}: data synchronization, segmentation and normalization.\par

For data synchronization, the ground truth from the Optojump system only provides start points of the stance and swing phases, in the unit of seconds, but the network input from  the sensor is samples from 1000Hz or 20Hz sampling rate. Thus, we synchronize the sensor data and the ground truth data according to the sampling rate of the IMU sensor, as shown in Fig. \ref{fig.Aftersyn}.  \par

After synchronization, we split the raw data to 1000-sample-point (1 second) windows with 500-sample-point (0.5 seconds) overlapping between windows, both in sensor data and ground truth. In which, one second window is chosen because one step is usually smaller than one second. Thus, this window can cover a whole step. In addition, the overlapping can make the model more robust across the window boundary. With this method, the whole dataset becomes 2604 seconds long, 1.6 times than the original one. Note that the testing data is also split to 1000-sample-point but without any overlapping.\par

 \begin{figure}[b]
    \includegraphics[width=\linewidth]{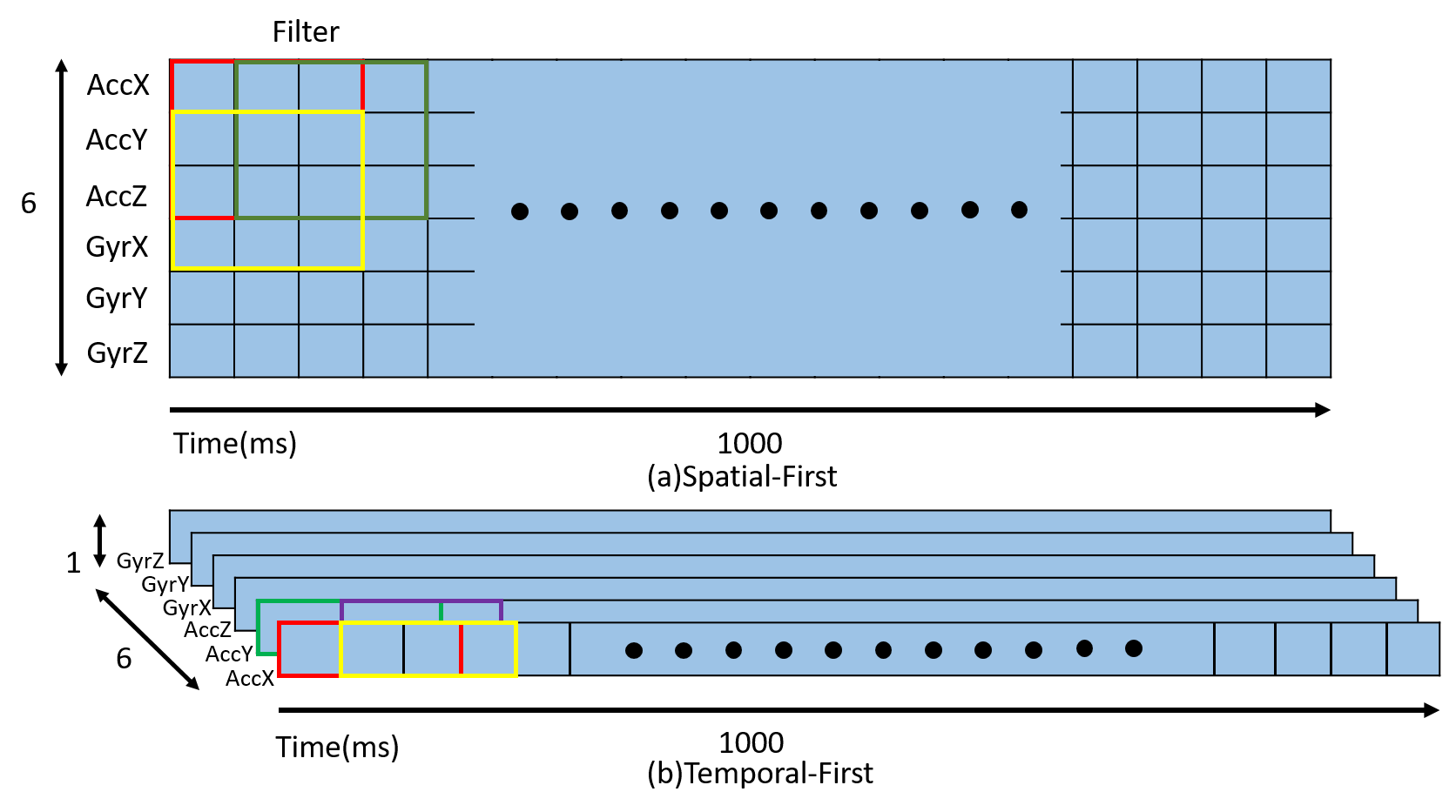}
    \caption{Two types of the IMU oriented order (a) for \emph{Spatial-First} and for \emph{Temporal-First}.} 
    \label{fig.sensororder}
\end{figure}

 The order of image segmentation input usually use \emph{CHW} (channel × height × width). But the time series data is expanded across spatial domain (different sensors) and temporal domain (different sample points). Fig. \ref{fig.sensororder} shows two types of IMU oriented processing order: \emph{Spatial-First} or \emph{Temporal-First}. The \emph{Spatial-First} input will arrange input data as 1 × 6 × 1000 (channel × sensors × one segment length) as shown in Fig. \ref{fig.sensororder} (a). By analogy with segmentation input, applying convolution on the \emph{Spatial-First} input will find correlation between different sensors first and then fuse different temporal features in the later layers. The \emph{Temporal-First} will arrange input data as 6 × 1 × 1000
(sensors × channel × one segment length), where 6 for 6-axis
sensor data (tri-axis accelerometer, and tri-axis gyroscope) and
1000 for data length in a window. Applying convolution on
the \emph{Temporal-First} input will find temporal correlation within each sensor first and then fuse different sensor features in the later layers. 
 

Following segmentation, we apply normalization on our dataset. In our data, the scale of accelerometer range and gyroscope range is different, ± 16 g (g = 9.81 m/$s^{2}$) and ± 2000 °/s separately. Therefore, we divide these sensor data by their maximum value to avoid imbalanced input data scale.\par

\subsection{Network Architecture}
\begin{figure}[tb]
    \includegraphics[width=\linewidth]{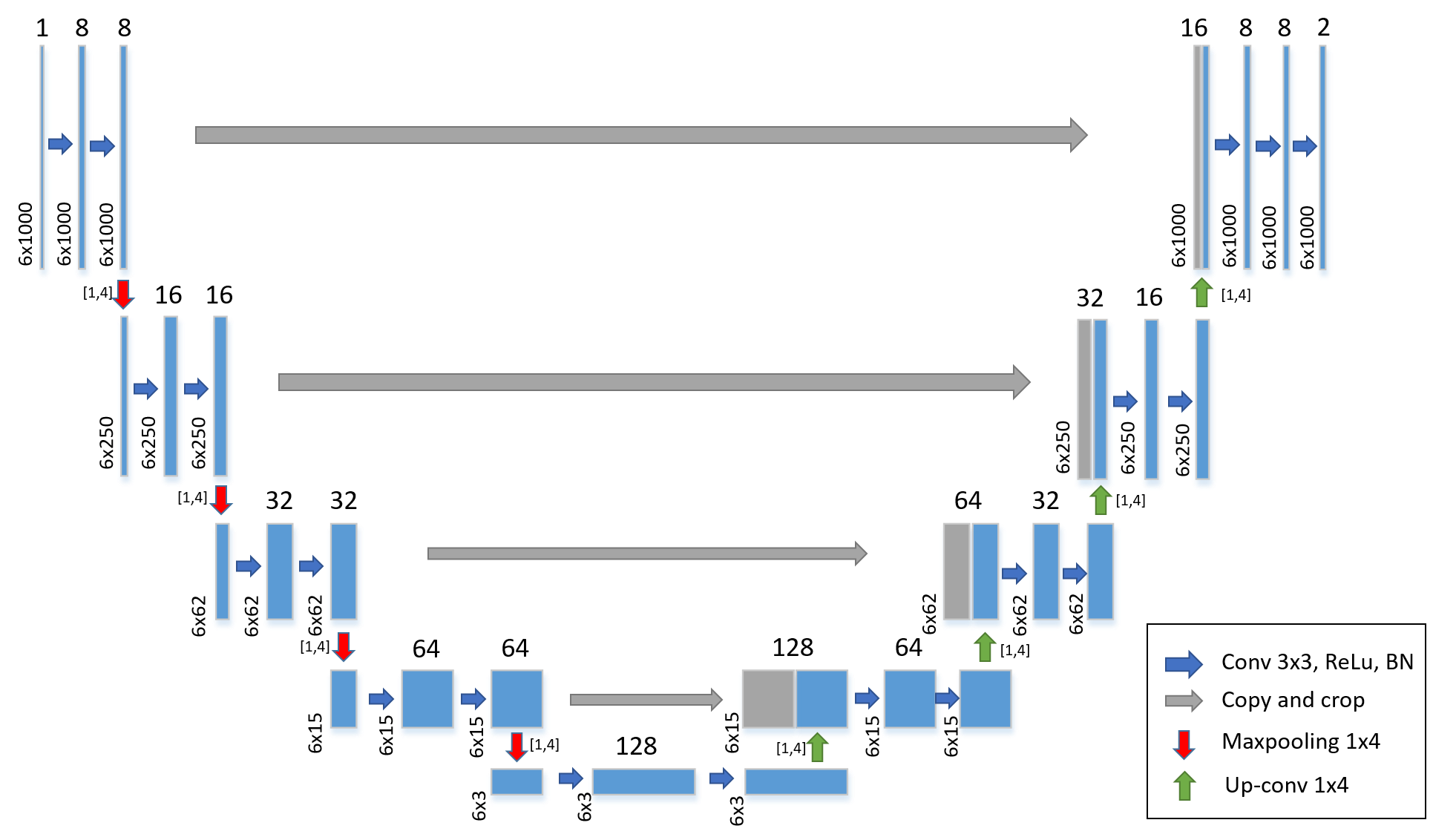}
    \caption{The structure of the \emph{IMU-Net}.} 
    \label{fig.IMUNET}
\end{figure}

The proposed network is based on \emph{U-Net} as shown in Fig. \ref{fig.IMUNET}. A naïve \emph{U-Net} application to this task will be like the following with some modifications. \emph{U-Net} contains two paths, encoding path (left side) and a decoding path (right side). The encoding path consists of several combinations of double-convolutional layers and max pooling layers. The double convolutional layer has two convolutional layers, and each convolutional layer uses 3 × 3 filters, rectified linear unit (ReLU) for activation function and batch-normalization. Following each double-convolutional layer is a 1 × 2 max-pooling layer instead of widely used 2 × 2 max-pooling. This max-pooling is to downsample data along the temporal domain instead of different axises of the sensor that will lose too much information. The decoding path is just the reverse of the encoding path. In which, each step concatenates upsampled feature maps with the correspondingly feature maps from the encoding path. Similar to the encoder side, the decoder uses 1 × 2 up-convolution. At the final layer, a convolutional layer is used to map feature maps into two classes, swing phase and stance phase. This model setting is denoted as \emph{U-Net} in this paper.\par

Above network has two drawbacks: small receptive field that will result in classification error, and high complexity that is not suitable for mobile devices. Fig. \ref{fig.IMUNET} shows the proposed network architectures that is modified from \emph{U-Net}, denoted as \emph{IMU-Net}. In the collected dataset, the stance time or swing time is usually larger than the half of input. However, with 1 × 2 max-pooling in \emph{U-Net},   the receptive field of \emph{U-Net} is 2 × 2 × 2 × 2 × 3 = 48, which is much smaller than half of input size, 500. The receptive field refers to the part of the input that is visible or effective in the last layer. To fit the receptive requirement of gait phase detection without complexity increase,  we modify the filter size of max-pooling layer from 1 × 2 into 1 × 4. Thus, the receptive field of \emph{IMU-net} is 4 × 4 × 4 × 4 × 3 = 768, larger than half of input size. To further reduce complexity, we also decrease the number of channels in each convolutional layer. Table \ref{tab.struc} shows the detailed network architectures of both models. The proposed model only needs 487K parameters, which is only 1.6\%  of parameters in \emph{U-Net}.

\renewcommand\arraystretch{1.37}
\begin{table}[tb]
\centering
\caption{The detail of both model.}
\scalebox{1}{
\begin{tabular}{|c|c|c|c|c|}
\hline
\multicolumn{3}{|c|}{Model}               & \emph{U-Net}          & \emph{IMU-Net}             \\ \hline
Side                 & Num. & Layers      & Channel/filter size & Channel/filter size \\ \hline
\multirow{9}{*}{Enc} & 1    & double-Conv & 64/3 × 3            & 8/3 × 3             \\ \cline{2-5} 
                     &      & Maxpooling  & */1 × 2             & */1 × 4             \\ \cline{2-5} 
                     & 2    & double-Conv & 128/3 × 3           & 16/3 × 3            \\ \cline{2-5} 
                     &      & Maxpooling  & */1 × 2             & */1 × 4             \\ \cline{2-5} 
                     & 3    & double-Conv & 256/3 × 3           & 32/3 × 3            \\ \cline{2-5} 
                     &      & Maxpooling  & */1 × 2             & */1 × 4             \\ \cline{2-5} 
                     & 4    & double-Conv & 512/3 × 3           & 64/3 × 3            \\ \cline{2-5} 
                     &      & Maxpooling  & */1 × 2             & */1 × 4             \\ \cline{2-5} 
                     & 5    & double-Conv & 1024/3 × 3          & 128/3 × 3           \\ \hline
\multirow{9}{*}{Dec} &      & Up-conv     & */1 × 2             & */1 × 4             \\ \cline{2-5} 
                     & 6    & double-Conv & 512/3 × 3           & 64/3 × 3            \\ \cline{2-5} 
                     &      & Up-conv     & */1 × 2             & */1 × 4             \\ \cline{2-5} 
                     & 7    & double-Conv & 256/3 × 3           & 32/3 × 3            \\ \cline{2-5} 
                     &      & Up-conv     & */1 × 2             & */1 × 4             \\ \cline{2-5} 
                     & 8    & double-Conv & 128/3 × 3           & 16/3 × 3            \\ \cline{2-5} 
                     &      & Up-conv     & */1 × 2             & */1 × 4             \\ \cline{2-5} 
                     & 9    & double-Conv & 64/3 × 3            & 8/3 × 3             \\ \cline{2-5} 
                     & 10   & Conv        & 2/3 × 3             & 2/3 × 3             \\ \hline
\multicolumn{3}{|c|}{Parameters}          & 29,649,794          & 487,154             \\ \hline
\multicolumn{3}{|c|}{FLOPs}               & 21.6G               & 0.78G               \\ \hline
\end{tabular}
}
\label{tab.struc}
\end{table}
\section{Training and evaluation}
Both models use cross entropy as the loss function. For model training, we use random subsets of the training dataset (mini-batches), and train the model for 500 epochs with batch size 100. For optimization, we use the Adam optimizer with default setting of $\beta$ = (0.9,0.999) and $\epsilon$ = 1e-8, set learning rate to 0.01, and initialize model parameters with random assignments from the normal distribution. For validation purposes, we applied a 3-fold cross validation with independent subjects on the training dataset and testing dataset. We also use 10\% of training dataset for validation dataset. Besides, for obtaining the best model, we monitor the loss of validation set in the training process and save the best model when validation loss is minimum.

\begin{table}[tb]
\centering
\caption{The performance of \emph{U-Net} and \emph{IMU-Net}, \emph{Temporal-First} and \emph{Spatial-First}, where the number in swing and stance time is mean ± standard deviation. }
\resizebox{\linewidth}{!}{
\begin{tabular}{ccccc}
\toprule
Model                                                         & \multicolumn{2}{c}{\emph{U-Net}} & \multicolumn{2}{c}{\emph{IMU-Net}}   \\ \hline
Type                                                          & \emph{Spatial-First}  & \emph{Temporal-First} & \textbf{\emph{Spatial-First}} & \emph{Temporal-First} \\ \hline
Swing(ms)                                                     & 15.41±15.68    & 28.23±38.15   & \textbf{9.80±10.47}    & 19.90±18.15   \\ \hline
Stance(ms)                                                    & 14.87±14.74    & 14.50±19.21   & \textbf{8.20±9.57}     & 10.43±12.27   \\ \hline
Gait phase accuracy(\%)                                                  & 95.98          & 95.44         & \textbf{96.88}        & 96.47         \\ \hline
\begin{tabular}[c]{@{}c@{}}Stride accuracy(\%)\end{tabular} & 98.69         & 97.17         & \textbf{99.59}        & 98.28         \\ \bottomrule
\end{tabular}}
\label{tab.result}
\end{table}

\section{Result and real-time implementation}

\subsection{Result}
Table \ref{tab.result} shows the error analysis of both \emph{U-Net} and \emph{IMU- Net} on different types of input data with four metrics. The four metrics are all evaluated according to per sample prediction result due to the segmentation approach. The accuracy of stance and swing time is represented by their mean error and standard deviation. The accuracy of gait phase is to divide the correct gait phase prediction points with whole number of sample points. Stride accuracy is to divide the correctly predicted stride count with the total number of strides, which is more coarse grained than the gait phase accuracy. \par

The result shows that the proposed \emph{IMU-Net} has better accuracy than the naïve \emph{U-Net} due to the gait phase aware receptive field setting to cover more global features. The mean error of stance and swing phase is both lower than 10ms. The \emph{Spatial-First} processing order has higher performance than other cases since the correlation between different sensors can be well extracted first by the convolution layer. Table \ref{tab.runwalkresult} shows that the \emph{IMU-Net} can get high performance on both walking and running case.\par

Table \ref{tab.compareresult} shows comparisons with other approaches. It is difficult to compare our result with other work due to different metrics, settings and datasets. However, with the segmentation approach, our approach can achieve much lower errors than others.

\begin{table}[tb]
\centering
\caption{The performance of two model for running and walking case. In \emph{Spatial-First} type.}
\resizebox{\linewidth}{!}{
\begin{tabular}{ccccc}
\toprule
Model             & \multicolumn{2}{c}{\emph{U-Net}} & \multicolumn{2}{c}{\emph{IMU-Nnet}} \\ \hline
Type              & Running        & Walking       & Running      & Walking       \\ \hline
Swing(ms)         & 15.36±13.95    & 15.82±27.11   & \textbf{9.26±9.57}    & \textbf{12.94±15.24}   \\
Stance(ms)        & 15.32±13.83    & 11.67±21.51   & \textbf{7.84±8.75 }   & \textbf{10.30±13.57}   \\
Accuracy(\%)      & 96.06          & 96.28         & \textbf{97.57}        & \textbf{97.52}         \\
Gait accuracy(\%) & 98.69          & 98.64         & \textbf{99.58 }       & \textbf{99.65}         \\ \bottomrule
\end{tabular}}
\label{tab.runwalkresult}
\end{table}

\begin{table}[tb]
\centering
\renewcommand\arraystretch{1.2}
\caption{The comparison of previous studies and our method.}
\resizebox{8cm}{!}{
\begin{tabular}{ccl}
\toprule
Author                  & Stance time error & \multicolumn{1}{c}{Swing time error} \\ \hline
Fabio A. Storm {[}21{]} & 31±10ms           & None                                 \\ \hline
J.-Y. Jung {[}15{]}     & \multicolumn{2}{c}{21.6ms (Continuous error)}            \\ \hline
R. L. Evans {[}14{]}    & \multicolumn{2}{c}{23ms (mean error)}                                 \\ \hline
\textbf{Our method }             & \textbf{9.80±10.47ms}      & \textbf{8.20±9.57ms}                          \\ 

\toprule
Author                  & \multicolumn{2}{c}{Stride detection accuracy}              \\ \hline
L. V. Nguyen [12]            & \multicolumn{2}{c}{97.2\%}                               \\ \hline
T. Steinmetzer [16]            & \multicolumn{2}{c}{95.8\%}                               \\ \hline
\textbf{Our method}            & \multicolumn{2}{c}{\textbf{99.59\%}} \\   

\toprule
Author                  & \multicolumn{2}{c}{Gait phase accuracy}  \\ \hline
Z. Ding [17]            & \multicolumn{2}{c}{96.4\%}               \\ \hline
\textbf{Our method}     & \multicolumn{2}{c}{\textbf{96.88\%}} \\ 

\bottomrule
\end{tabular}}

\label{tab.compareresult}
\end{table}

\subsection{Real time implementation with low sampling rate}
The proposed model has implemented on mobile phone with real time data connection from sensors. However, the sampling rate of the sensor is constrained to 20 Hz in this case. Thus, the model is retrained with 20Hz sampled data. We downsample the dataset from 1000 Hz to 20 Hz by random downsampling, which is repeated 20 times to increase data amount to compensate the loss due to lower sampling rate. Thus, the dataset becomes 52080 seconds with 20 sample points per second. We use the 20 Hz dataset as the training dataset, modify the pooling size to 1 × 2 and apply the same training procedure. Table \ref{tab.20Hzresult} show the 3-fold cross validation result, which has even slightly higher performance than the original one due to data augmentation.\par

\begin{table}[tb]
\centering

\caption{The accuracy of \emph{IMU-Net} implement on 20Hz.}
\resizebox{\linewidth}{!}{
\begin{tabular}{cccc}
\toprule
\multicolumn{4}{c}{\emph{IMU-Net} @ 20Hz}                         \\\hline
Swing(ms)  & Stance(ms) & Gait phase accuracy(\%) & Stride accuracy(\%) \\
8.86±11.62 & 9.12±12.05 & 96.44        & 99.97   \\
\bottomrule
\end{tabular}}
\label{tab.20Hzresult}
\end{table}

The trained model is deployed as Android application with TensorFlow Lite.  Table \ref{tab.deviceresult} shows the detailed results on multiple phones.  Note that the execution time is the time of infering our model single time (one second sensor data) on the device. Based on the result, the maximum execution time is 36 ms, which is 3.6\% of one second, and achieves real-time execution.\par

\begin{table}[tb]
\renewcommand\arraystretch{1.2}
\centering
\caption{The execution time on different device.}
\resizebox{\linewidth}{!}{
\begin{tabular}{cccc}
\toprule
Device            & OS        & CPU                     & \begin{tabular}[c]{@{}c@{}}Execution \\     Time(ms)\end{tabular} \\ \hline
Samsung Galaxy J5 & Android 5 & Qualcomm Snapdragon 410 & 36                                                                 \\
Redmi Note 5      & Android 9 & Qualcomm Snapdragon 636 & 26                                                                 \\
HTC U11           & Android 7 & Qualcomm Snapdragon 835 & 18                                                                 \\
Google Pixel 3 XL & Android 9 & Qualcomm Snapdragon 845 & 9                                                                 \\
Asus Zenfone 3    & Android 8 & Qualcomm Snapdragon 625 & 20                                                                 \\
Asus Zenfone 6    & Android 9 & Qualcomm Snapdragon 855 & 18                                                                 \\ \bottomrule
\end{tabular}}
\label{tab.deviceresult}
\end{table}

\section{Conclusion}
This paper has presented a segmentation network approach for accurate gait phase detection based on single IMU sensor data. This deep learning approach enables end-to-end training without explicit calibration or design assumption, which successfully adapts to various speed of walking and running. The network based on \emph{U-Net} has modified with gait phase aware receptive field settings and IMU oriented processing order. With 20 Hz sampling rate, the model achieves average error of 8.86 ± 11.62 ms on swing phase, 9.12 ± 12.05 on stance phase, 96.44 \% accuracy of phase detection and 99.97 \% accuracy of stride detection. The model can run also on mobile phones in real-time with low complexity.\par


\end{document}